\documentclass[letterpaper, 10 pt, conference]{ieeeconf}  % Comment this line out if you need a4paper

\IEEEoverridecommandlockouts                              % This command is only needed if 
\usepackage{amsmath} % assumes amsmath package installed
\usepackage{amsfonts}

\usepackage{graphicx}
\usepackage{bm}
\usepackage{dblfloatfix}
\usepackage{cite}

\title{\LARGE \bf
Generating robotic elliptical excisions with human-like tool-tissue interactions}

\author{Artūras Straižys$^{1}$, Michael Burke$^{1,2}$ and Subramanian Ramamoorthy$^{1,3}$% <-this % stops a space
\thanks{$^{1}$School of Informatics, University of Edinburgh}%
\thanks{$^{2}$Department of Electrical and Computer Systems Engineering, Monash University}%
\thanks{$^{3}$Work supported by a grant from the UKRI Strategic Priorities Fund to the UKRI Research Node on Trustworthy Autonomous Systems Governance
and Regulation (EP/V026607/1, 2020-2024).}
\thanks{For the purpose of open access, the author(s) has applied a Creative Commons Attribution (CC BY) license to any Accepted Manuscript version arising.}
}

\begin{document}

\maketitle
\thispagestyle{empty}
\pagestyle{empty}

%%%%%%%%%%%%%%%%%%%%%%%%%%%%%%%%%%%%%%%%%%%%%%%%%%%%%%%%%%%%%%%%%%%%%%%%%%%%%%%%
\begin{abstract}
In surgery, the application of appropriate force levels is critical for the success and safety of a given procedure.
While many studies are focused on measuring in situ forces, little attention has been devoted to relating these observed forces to surgical techniques. Answering questions like “Can certain changes to a surgical technique result in lower forces and increased safety margins?” could lead to improved surgical practice, and importantly, patient outcomes. However, such studies would require a large number of trials and professional surgeons, which is generally impractical to arrange. Instead, we show how robots can learn several variations of a surgical technique from a smaller number of surgical demonstrations and interpolate learnt behaviour via a parameterised skill model. This enables a large number of trials to be performed by a robotic system and the analysis of surgical techniques and their downstream effects on tissue. Here, we introduce a parameterised model of the elliptical excision skill and apply a Bayesian optimisation scheme to optimise the excision behaviour with respect to expert ratings, as well as individual characteristics of excision forces. Results show that the proposed framework can successfully align the generated robot behaviour with subjects across varying levels of proficiency in terms of excision forces.

\end{abstract}

%%%%%%%%%%%%%%%%%%%%%%%%%%%%%%%%%%%%%%%%%%%%%%%%%%%%%%%%%%%%%%%%%%%%%%%%%%%%%%%%
\section{Introduction} \label{intro}
Surgical excision implies the application of physical forces necessary for tissue separation \cite{golahmadi2021tool}. A successful procedure requires appropriate levels of excision forces - sufficient for cutting, yet conservative to avoid damaging tissue (excessive force can account for more than half of the medical errors committed by surgical trainees \cite{tang2005analysis}). Cutting forces, on the other hand, strongly depend on the configuration of the blade \cite{atkins2004cutting, liu2021advance,liu2022recent}, and therefore on the excision technique. In order to derive optimal surgical techniques, in the context of surgical training or autonomous surgery, tool-tissue interaction forces and their downstream effect on the tissues must be studied in a controlled manner. A comprehensive analysis of a wide range of behaviours across different levels of expertise is needed to identify good and bad practices and to measure their benefit or harm. 

Unfortunately, such studies require a large number of participants at different stages of their professional development, which is extremely time-consuming and challenging in terms of logistics. As an alternative, we can use a considerably smaller number of subjects to teach a robot performing the excisions from various levels of proficiency and apply machine learning techniques to interpolate the behaviour between the demonstrations. A suitably parameterised behaviour model would let us generate a large number of trials with tight control over the process parameters, such as excision velocity, blade insertion angle, etc. Such a robotic setup would facilitate the exploration of the downstream effects of the cutting technique on the tissues, and therefore provide a deeper insight into its efficacy and safety. Importantly, this gives us the ability to align robotic cutting behaviour with the desired characteristics of excision forces to analyse various techniques and their influence on tissue outcome.

A standard approach to accomplish this relies on imitation learning techniques \cite{hussein2017imitation}, such as the Dynamic Movement Primitive (DMP) \cite{ijspeert2013dynamical}, to learn excision behaviours directly from demonstrations. In the case of the classical DMP formulation, the encoded policy can be generalised via hyper-parameters, such as the goal state and temporal scaling, as well as the coupling terms \cite{hoffmann2009biologically}. 
Unfortunately, although this allows exploration around the demonstrated trajectories, the parameterised policies are restricted to individual demonstrations with no relation to one another. 
Alternatively, one could apply DMP formulations that leverage multiple demonstrations to capture behaviour variability as a separate task parameter \cite{matsubara2010learning,zhao2014generating}, which can be interpolated to synthesise unseen behaviours \cite{matsubara2011learning}. In the context of learning surgical excisions, these methods allow explicit capture of different ``styles'' of cutting tissues, which is particularly relevant for our task. However, these methods are unsuitable to encode the end-effector’s pose trajectory in Cartesian space, as the orientation component requires special treatment of the SO(3) structure \cite{ude2014orientation}. This presents challenges to apply the above methods to learn cutting skills, as the position and orientation trajectories of a blade are inherently coupled due to the nonholonomic nature of the cutting motion \cite{rahal2019haptic,straivzys2023learning}.

\begin{figure*}[t]
\centering
\includegraphics[width=0.9\textwidth]{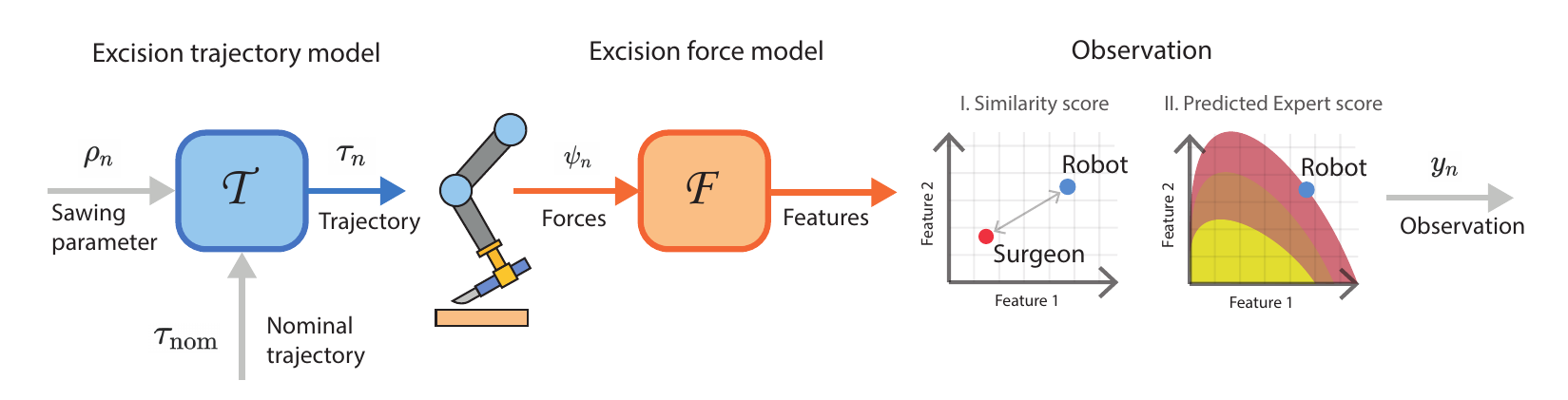}
\caption{The proposed framework to generate human-like excision behaviours operates as follows. At a high level, an objective function takes in the sawing parameter $\rho$ and outputs an observation similarity score $y$. The goal of the optimiser is to find the value for $\rho$ that maximises the observation variable $y$. The inner structure of the proposed objective function consists of the proposed trajectory generator ($\mathcal{T}$) that given the scalar $\rho \in [1,10]$ and a nominal trajectory $\tau_{\text{nom}}$, generates the pose trajectory of the blade $\tau$. Next, the robot executes the trajectory $\tau$, and collected forces $\psi$ are converted to a set of performance features by model $\mathcal{F}$. The obtained features are then used to define the final objective function, e.g. the similarity score between the robot-executed task and the performance of an actual surgeon, which updates the optimiser and generates a new excision behaviour. }
\label{fig:objective_fun}
\end{figure*}

As an alternative, in this paper we introduce a simple parametric model of elliptical excision that decomposes the skill into nominal and behaviour-driven components, each of which can be learned from demonstrations. Here, we focus on modelling a sawing movement as the most dominant characteristic of the elliptical excision technique \cite{rohrer2014surgery} observed in human trials \cite{straivzys2023generative}. Our model encodes the behaviour with a single real-valued parameter $\rho$ that determines the amount of sawing movement applied during excision. We then show how this model can generate a variety of human-like excision trajectories, and apply Bayesian optimisation to align the generated behaviour with expert ratings. 

Finally, as the core contribution of this paper, we propose a framework for aligning human-like robotic elliptical excisions with the desired characteristics of excision forces and demonstrate its applicability for analysing excision skills.
In this framework (Fig. \ref{fig:objective_fun}), we generate an elliptical excision behaviour using the parametric model described above, which generates the pose trajectory of the blade, specified by a behaviour parameter $\rho$ and a nominal cutting trajectory. The robot executes the generated trajectory in a real-world experiment. A suitable model characterising the excision behaviour from the excision force measurements \cite{straivzys2023generative,baghdadi2023tool,trejos2014development} allows us to 1) calibrate the robotic behaviours to match the desired characteristics of tool-tissue interaction, and 2) use the aligned robotic behaviour to analyse the excision techniques in a well-controlled and repeatable manner. Here, we propose a Bayesian Optimisation scheme to minimise the number of phantom excisions required for behaviour tuning.

% In summary, this paper presents the following contributions:
% \begin{itemize}
%     \item A generative model of blade motion for learning and reproducing human-like behaviours in the elliptical excision task. 
%     \item A novel framework for analysing surgical excision techniques and skills using a robotic reproduction of various cutting behaviours with human-like characteristics of excision forces. 
% \end{itemize}

% We organise the paper as follows. First, we introduce the parametric generator of elliptical excision trajectories. Next, we briefly review one of the approaches for characterising excision performance from force measurements, suitable for the proposed framework. We then describe the proposed method for aligning the generated excision behaviours with respect to the desired excision characteristics. Finally, we illustrate the applicability of the proposed framework with a set of practical excision experiments and discuss our results and findings.

\begin{figure*}[t]
\centering
\includegraphics[width=0.9\textwidth]{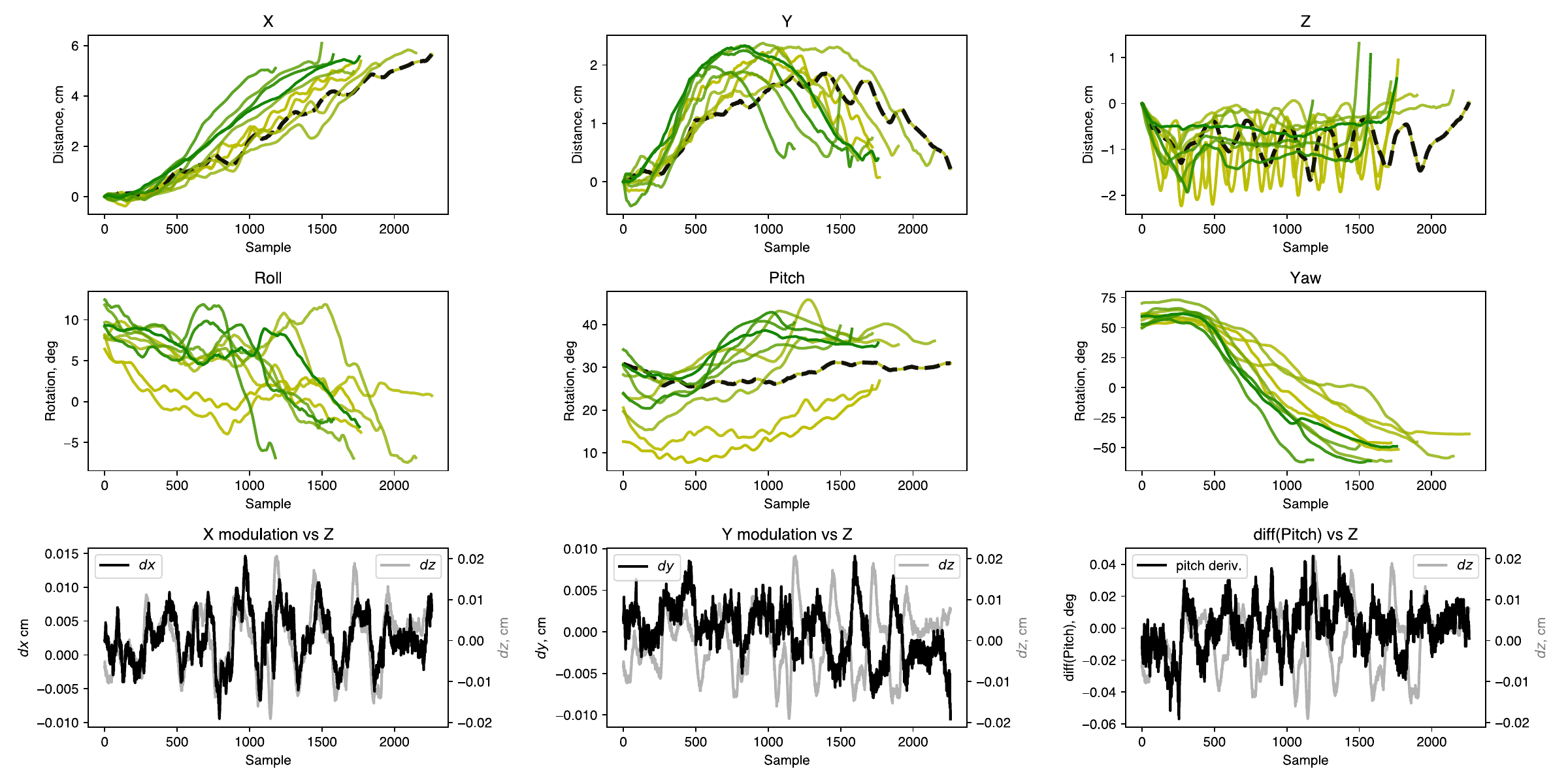}
\caption{ Measured individual position (top row) and orientation (middle row) trajectories of the blade for each of the demonstrated behaviour (the darker green lines correspond to smoother excisions). (Bottom row) Differenced measurements of $x$, $y$ and pitch versus differenced measurements of $z$. Note: the original measured trajectories are shown as black dashed lines in the above plots. }
\label{fig:measured_traj}
\end{figure*}

\section{Parametric generation of human-like excision trajectories} \label{trajectory-model}

The primary objective of our excision trajectory generator is to produce realistic pose trajectories of the blade, mirroring those typically observed during an elliptical excision procedure. In particular, we are interested in learning the auxiliary sawing movement of the blade, which can play a role in assisting the excision by lowering the cutting forces required for the task \cite{atkins2004cutting}. In this section, we introduce a generative model of blade trajectories with a single parameter $\rho$ that specifies the amount of sawing movement applied to any desired elliptical excision motion. 

Our central hypothesis is that elliptical excision comprises two movement components: a nominal smooth cutting motion along the ellipse and an adaptive sawing motion that assists the excision. To better understand the interplay between these two components, we recorded ten elliptical excisions with varying amounts of sawing behaviour, from a highly pronounced sawing movement to an extremely smooth excision. Fig. \ref{fig:measured_traj} (top and middle rows) shows the measured position and orientation trajectories of the blade in each trial. Notice the back-and-forth oscillation of the blade on the $XY$ plane, a distinct feature of the sawing movement. It should be noted that sawing is executed at a noticeably different pitch when compared to smoother executions. Unsurprisingly, smoother excisions result in a much lower spread of the pose trajectories, which highlights the challenge of maintaining consistent motion when sawing.
As expected, the sawing motion is most dominant along the $z$ axis, i.e. the cutting depth. However, the sawing is also reflected in the blade's $x$, $y$ and pitch trajectories. 

\begin{figure*}[t]
\centering
\includegraphics[width=0.9\textwidth]{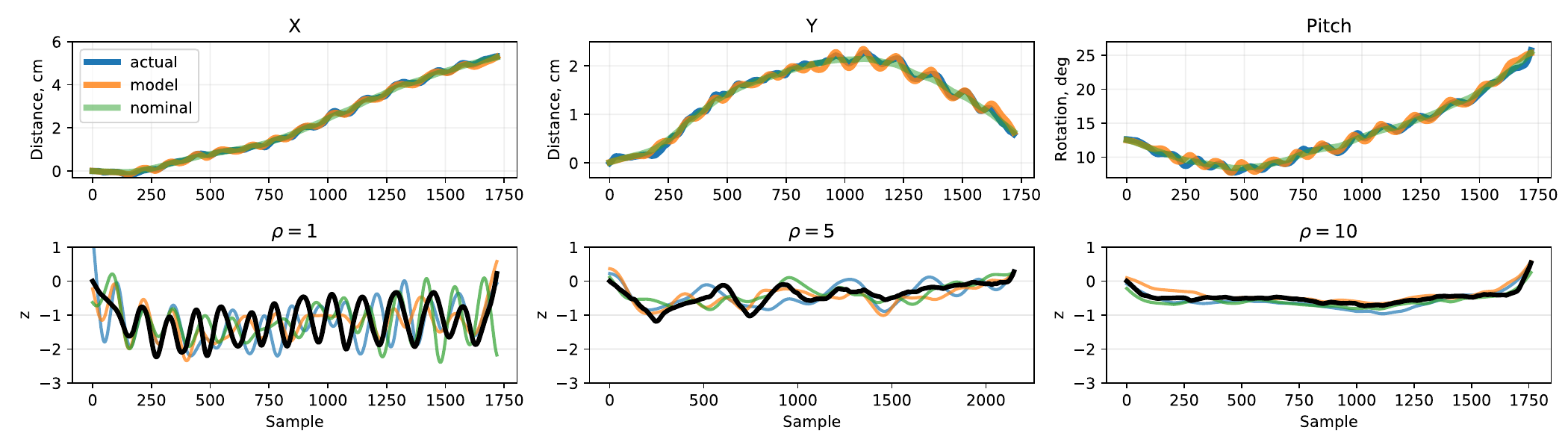}
\caption{(Top row) Comparison of measured $x$, $y$ and pitch trajectories (blue lines) with model predictions (orange lines). The nominal trajectories obtained by low-pass filtering raw measurements are denoted with green. (Bottom row) Comparison of measured $z$ trajectories for three different sawing behaviours (black lines) and corresponding synthetic trajectories generated by the model (coloured semi-transparent lines). }
\label{fig:model_pred}
\end{figure*}

Fig. \ref{fig:measured_traj} (bottom row) shows the trajectories of differenced $x$, $y$ and pitch motion components, compared to a differenced $z$ trace, all from one of the observed sawing behaviours. The entire trace of $x$, and the first half of the parabola in $y$ trajectory, are noticeably in-phase with the $z$. In contrast, the whole trace of the pitch, and the second half of the parabola in $y$ trajectory, are noticeably out-of-phase with the $z$ trajectory. The relationship between $z$ and $xy$ trajectories indicates that sawing movement is achieved by propagating the blade forward on the ascent, with residual backward motion on the blade's descent. The relation between $z$ and pitch trajectories suggests that consistent modulation of the insertion angle is also a part of the slicing motion. 

\subsection{Modeling blade trajectories for elliptical excisions}
Given the above insights, we model the elliptical excision behaviour as follows. First, we decompose the cutting motion into a \textit{nominal} movement component that follows the desired smooth cutting contour, and a \textit{behaviour} component that characterises the manner of task execution (e.g. sawing vs smooth excisions). For the $z$ trajectory, this is written as
\begin{equation}
    z_m = z_{n} + z_{b}
\end{equation}
where $z_m$, $z_n$ and $z_b$ are the measured, nominal and behaviour trajectories along the $z$ axis, respectively. 

Next, we assume that a nominal movement component can be approximated by low-pass filtering of the measurements, and therefore the behaviour component can be computed as $z_b = z_m - z_n$. The obtained behaviour component $z_b$ can be used to model the observed $x$, $y$ and pitch trajectories of the blade, as follows:
\begin{equation} \label{model}
\begin{array}{lcl}
     & x_m   = x_n + c_x z_{b} \\
     & y_m   = y_n + c_y z_{b} \\
     & \beta_m   = \beta_n + c_\beta z_{b}
\end{array}
\end{equation}
Here $x_m$, $y_m$ and $\beta_m$ are the measured $x$, $y$ and pitch trajectories, respectively, and $x_n$, $y_n$ and $\beta_n$ are their corresponding nominal trajectories. $c_x$, $c_y$ and $c_\beta$ are the scaling coefficients (We invert the sign of $c_y$ in the second half of the $y_m$ parabola to correctly represent the modulation of parabolic nominal trajectories on the $XY$ plane).

As a result, we model the $x$, $y$ and pitch trajectories as a function of $z_b$. According to this model, any modulation along the cutting depth is reflected in the $x$, $y$ and pitch trajectories. Fig. \ref{fig:model_pred} (top row) compares the actual trajectories with those predicted by our model for one of the observed sawing behaviours. Here, we used a first-order Butterworth filter with a cutoff frequency of 0.6 Hz to obtain the nominal trajectories from the raw measurements and set $c_x$, $c_y$ and $c_{\beta}$ parameters to 0.2, 0.13 and -1.2 values, respectively. 

\subsection{Learning cutting behaviour}
The movement component $z_b$, which defines the excision behaviour, can be learned using one of many supervised learning techniques. In this work, we first apply the following function approximation to encode the behaviour component $z_b$:
\begin{equation}
    z_b(t) \approx \frac{\sum_{i=1}^N \psi_i(t)\theta_i}{\sum_{i=1}^N \psi_i(t)},
\end{equation}
where $\psi_i(t) = \textrm{exp}(-h_i(t-c_i)^2)$ are Gaussian basis functions, $N$ is the number of basis functions, $t$ is the timestep, $c_i$ and $h_i$ are the centres and widths of the basis functions, and $\theta_i$ are the weights of the basis functions. 

The vector of learned weights $\bm{\theta} = [\theta_1, ..., \theta_N]$ can be viewed as a compressed representation of the $z_b$ time series. 
We fit a second-order autoregression model to vector $\bm{\theta}$:
\begin{equation} \label{autoreg}
    \theta_i = c + a_1 \theta_{i-1} + a_2 \theta_{i-2} + \epsilon_i
\end{equation}
where $a_k$ and $c$ are the autoregression coefficients and bias constant, respectively; $\epsilon_i$ is the white noise term.

Finally, given the excision behaviour label (1 to 10, where 1 represents the distinct sawing cuts, and 10 denotes smooth excisions), we fit a linear model to predict the $a_1$ and $a_2$ coefficients. Thus, given a real number $\rho \in [1,10]$, our model generates the behaviour component $z_b$, which along with a nominal cutting trajectory can be used to produce a human-like pose trajectory of the blade with desired sawing behaviour to be executed by a robot. Fig. \ref{fig:model_pred} (bottom row) compares the $z$ trajectories sampled from our model with the actual measurements $z_m$ for different sawing behaviours. 

\section{Optimisation of elliptical excision technique}

% The parametrisation of excision behaviour described above allows optimisation of the excision behaviour with respect to various performance characteristics. In this work, we focus on the tool-tissue interaction forces that arise during the excision procedure and are important to its efficacy and safety. 
Optimisation of the robotic behaviour with respect to the excision force characteristics can offer interesting prospects for studies on the effects of surgical procedures. For example, the alignment of robot behaviour with excision forces that match those of expert surgeons is particularly promising for robotic surgery applications. 
Equally important is the ability to faithfully recreate sub-standard excision techniques and study the common mistakes observed in less experienced surgeons or trainees.
In addition, technique optimisation concerning force characteristics critical to the procedure's safety could positively contribute to the existing surgical training and practice. 

In this work, we use an excision force model described in \cite{straivzys2023generative}, characterising the excision performance from force measurements; however, other force-based characterisation approaches can be readily applied in the proposed framework. 
Below, we provide a brief overview of the force-based characterisation of elliptical excision used in this study, followed by descriptions of the objective function and the proposed method for technique optimisation.

\subsection{Performance characterisation from force measurements}
A model of excision forces introduced in \cite{straivzys2023generative} allows characterising the performance of the generated excision behaviours directly from the measurements of the excision forces. Alongside descriptive statistics of excision forces, this model parameterises task-related characteristics correlated with elliptical excision performance, such as abruptness of task execution flow. This approach models the elliptical excision process as a hybrid system, with underlying continuous dynamics of viscoelastic interaction between tissues and the blade, as well as discrete event dynamics, typically associated with tissue re-tensioning or blade re-orientation. 

First, the model approximates the process of cutting a viscoelastic object as a continuous blade's movement through Maxwell material using the following constitutive law:
\begin{equation}\label{maxwell}
    \frac{\eta}{E}\dot{f} + f = \eta \dot{x}  
\end{equation}
with $f$ is the excision force, $\dot{f}$ the time derivative of the excision force, $\dot{x}$ is the blade's velocity,$E$ and $\eta$ the Maxwell model's spring and damper coefficients, respectively. 

The model assumes that an excision is executed using $K$ cutting regimes, where each regime $k$ corresponds to a constant velocity of the blade $v_k$. System uncertainty is modelled using white Gaussian noise with variance $\sigma_k$, $\Tilde{v}_k \sim \mathcal{N}(v_k,\sigma_k^2)$. Cutting regimes are switched according to $K \times K$ transition matrix $\mathbf{Q}$, which along with $v_k$ and $\sigma_k^2$, can be learned by fitting a Hidden Markov Model (HMM) to $\dot{x}$. Under the assumption of the Maxwell model, $v_k$ can be obtained directly from force measurements $f$, captured using a suitably instrumented scalpel. 

The learned model parameters encode the amplitude and temporal features of the excision forces that characterise the manner of task execution. For instance, parameters $\{v_1$...$v_K\}$ describe the dominant force levels of the excision forces, and thus the overall magnitude and spread of forces applied during the excision. Along with transition probability matrix $\mathbf{Q}$, which captures the temporal characteristics of the excision forces, these parameters can encode meaningful features, such as Energy (where increased energy reflects higher cutting forces applied for a longer duration) or Smoothness (where increased smoothness reflects the higher probability of sudden rise and fall of excision forces). 

\subsection{Objective function}
Fig. \ref{fig:objective_fun} shows a diagram of the proposed framework for aligning robotic cutting behaviour with desired characteristics of excision forces, typically observed in human trials. At every $n$-th iteration, the robot executes the excision trajectory $\tau_n$ generated by the proposed trajectory model ($\mathcal{T}$) using candidate behaviour parameter $\rho_n$ and a fixed nominal trajectory $\tau_{\textrm{nom}}$. After the execution, the recorded excision force profile $\psi_n$ is provided to the force-based characterisation model ($\mathcal{F}$), which encodes the characteristics of the excision forces, as described in the previous section. The features extracted by model $\mathcal{F}$ are then used to evaluate the objective function $g(\rho_n)$. Two scoring methods are used to evaluate the proposed objective function:

\subsubsection{A single characteristic of excision forces} is used for aligning the generated robot behaviour with respect to a certain characteristic of the excision forces, e.g. amplitude or smoothness evaluated by model $\mathcal{F}$.

\subsubsection{Expert rating} \label{objfun2}
This objective function is used to optimise the robot behaviour with respect to a predicted expert score. \cite{straivzys2023generative} evaluated the magnitude-based characteristics of the excision forces (the Amplitude and Consistency features, captured by model $\mathcal{F}$) using 15 expert-labelled excision trials. Here, we apply k-nearest neighbour regression (with $k=1$) to predict the expert rating given the excision features.

% Finally, the evaluated objective function $y_n = g(\rho_n)$, along with $\rho_n$ value are fed back to the optimiser for the next iteration. 

\begin{figure}[t]
\centering
\includegraphics[width=0.3\textwidth]{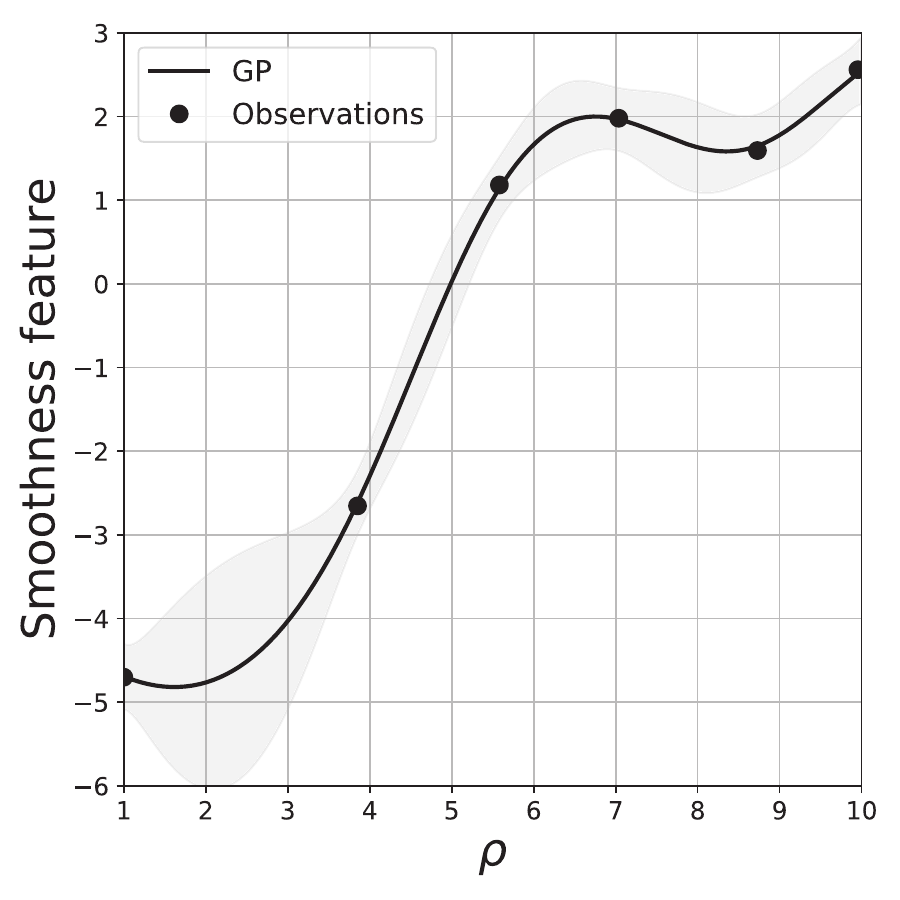}
\caption{Gaussian process model fit to six datapoints (black dots) collected by optimising $\rho$ with respect to the Smoothness feature. (The black line is the posterior mean, and the shaded region is 95$\%$ confidence interval).}
\label{fig:smoothness_feature}
\end{figure}

\subsection{Bayesian optimisation of excision behaviour}
% Below, we describe a method for such optimisations. 
% The purpose of optimisation is to obtain the sawing parameter $\rho$, that would maximise the experts' evaluation scores predicted by the excision force model, and to analyse how changes in the sawing behaviour affect the excision force characteristics. 
We apply Bayesian Optimisation (BO)\cite{frazier2018tutorial} to reduce the number of cutting experiments required to generate a given human-like excision behaviour. 
At each iteration, BO optimises an acquisition function $\alpha$ to choose the next candidate sawing parameter $\rho$ for trajectory generation in order to evaluate the objective function $g$ (as described above). We model the mapping beween trajectory parameter $\rho$ and the objective function using a Gaussian process (GP) \cite{rasmussen2005gaussian}:
\begin{equation}
    g(\rho) \sim \mathcal{GP}\left( m(\rho), \kappa(\rho, \rho') \right)
\end{equation}
where $m(\rho)$ and $\kappa(\rho, \rho')$ are the mean and kernel functions. 

\begin{figure*}[!t]
\centering
\includegraphics[width=0.9\textwidth]{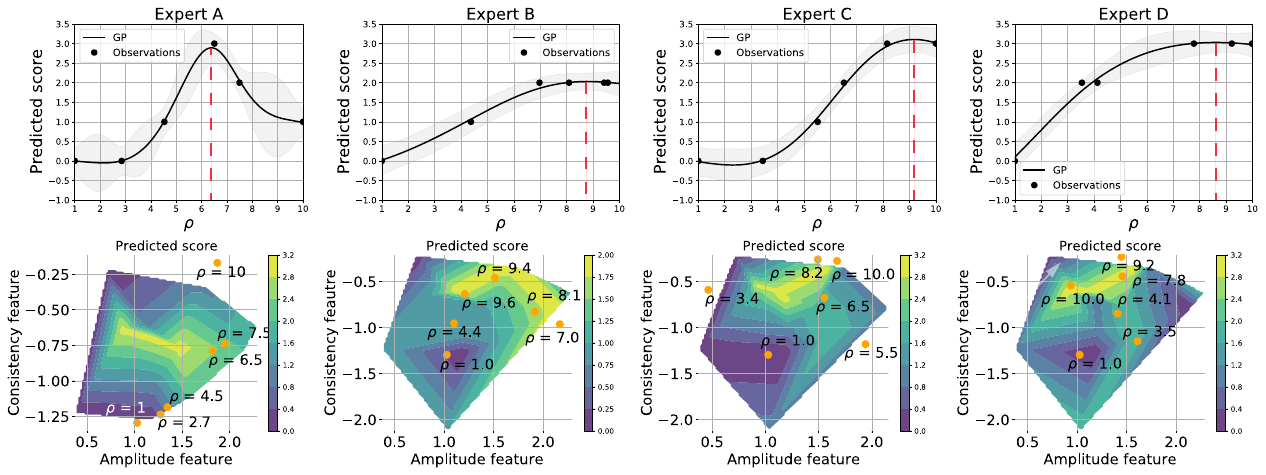}
\caption{Results for behaviour ($\rho$) optimisation with respect to the interpolated performance score from four experts. (Top row) Gaussian process models fit to six observations (black dots) obtained during optimisation. The black lines show the posterior, the shaded regions illustrate the 95$\%$ confidence intervals, and the red dashed lines highlight the optimal $\rho$ parameters for each of the experiments. (Bottom row) Contour plot of the interpolated expert scores over the feature space of the excision force model $\mathcal{F}$. The orange dots are the individual sample points $\rho$ used during optimisation. }
\label{fig:learning_experts}
\end{figure*}

For $\alpha$, we use the Expected Improvement acquisition function with the following closed-form expression \cite{jones1998efficient}:
\begin{equation}
\begin{array}{lcl}
    \alpha(\rho)= 
    \left( m(\rho) - g(\rho^+) - \epsilon \right) \Phi(Z) + \sigma(\rho)\phi(Z),
\end{array}
\end{equation}
where 
\begin{equation*}
    Z = 
\begin{cases}
    \frac{m(\rho)-g(\rho^+)-\epsilon}{\sigma(\rho)},& \text{if } \sigma(\rho) > 0\\
    0,              & \text{if } \sigma(\rho) = 0
\end{cases}
\end{equation*}
Above, $m(\rho)$ and $\sigma(\rho)$ are the mean and the standard deviation of the GP posterior, and $\Phi$ and $\phi$ are the cumulative probability function and the probability density function, respectively. $\rho^+$ is the current optimal choice of sawing parameter, and $\epsilon$ is a scalar that sets the tradeoff between exploration and exploitation during optimisation. 

We use a Mat\'ern kernel \cite{stein1999interpolation} function $\kappa(\rho,\rho')$ with the following analytical expression:
\begin{equation}
    \kappa(\rho_i,\rho_j) = \frac{1}{\Gamma(\nu) 2^{\nu-1}} \Big( \frac{\sqrt{2\nu}}{l}r \Big)^{\nu} K_{\nu} \Big( \frac{\sqrt{2\nu}}{l}r \Big)
\end{equation}
where $r = |\rho_i - \rho_j|$, $\Gamma(\cdot)$ is the gamma function, $K_\nu$ the Bessel function, $\nu$ and $l$ kernel hyperparameters. 

In our experiments, we set $\epsilon = 2$, $\nu = 2.5$ and $l = 1$. These parameters were chosen manually during setup calibration.

\section{Experiments and Results}
We performed two sets of experiments with the following objectives: 1) to find the excision technique that maximises the smoothness feature of the applied cutting forces, and 2) to find the excision technique whose excision forces predict the highest expert ratings.  

Fig. \ref{fig:smoothness_feature} shows the optimisation results for six iterations of smoothness feature optimisation. The first trial was initialised with a sample $\{ \rho_1 = 1, y_1 \}$, where $y_1$ is the force smoothness feature evaluated by model $\mathcal{F}$. Optimisation results confirmed our expectations that smoother excision trajectories (e.g. generated by model $\mathcal{T}$ using larger values of $\rho$ parameter) result in smoother excision forces.

\begin{figure*}[t]
\centering
\includegraphics[width=0.9\textwidth]{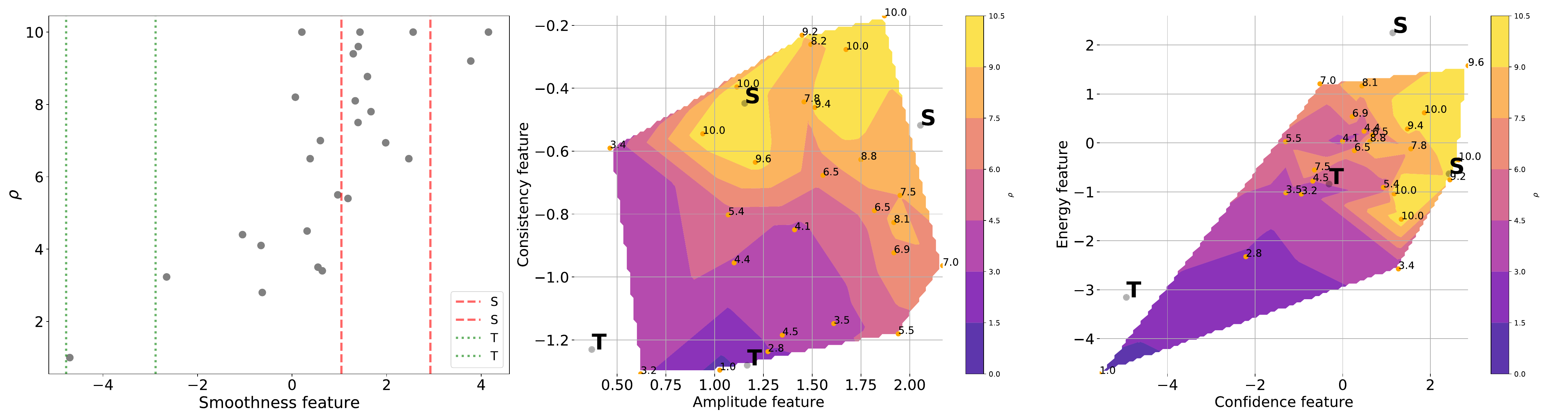}
\caption{ (Left) $\rho$ parameter vs evaluated smoothness feature. (Middle and right) Contour plots of the $\rho$ parameter values obtained during experiments. The orange dots are the individual datapoints with shown values for $\rho$ parameter. Note:  $\textbf{T}$ and $\textbf{S}$ denote medical trainee and professional surgeon, respectively. }
\label{fig:surgeons-trainees}
\end{figure*}

Fig. \ref{fig:learning_experts} shows the results for six iterations of expert score optimisation, for each of the four experts. As before, we initialised the optimisation with $\{ \rho_1 = 1, y_1 \}$ pair (this time, $y_1$ is the predicted expert score, as described in section \ref{objfun2}). The obtained mean of the posterior GP predicts higher expert scores for the excision behaviours generated using larger $\rho$ values (i.e. smoother trajectories). In other words, the results suggest that sawing movement is more likely to be penalised by experts. In addition, the experiment demonstrates that the sawing behaviour parameterised by $\rho$ can achieve different modulations of excision forces. For example, the score from Expert A highlights the rater's preference towards excisions with a more pronounced force modulation, as reflected by the Consistency feature. Notice that the posterior GP successfully captures this preference with the optimal sawing parameter $\rho \approx 6$. 

\subsection{Excision force characteristics versus $\rho$ parameter}
We analysed the $\{ \rho, y \}$ datapoints collected in the above experiments to explore the relationships between the sawing parameter $\rho$ and characteristics of the excision forces. Fig. \ref{fig:surgeons-trainees} (left) shows a scatter plot of the $\rho$ parameter values vs the smoothness feature of the excision forces evaluated by model $\mathcal{F}$. As in the first experiment, the sawing behaviour shows a strong positive correlation with the Smoothness feature (Pearson's $r=0.74$, $p < 0.05$). This relationship has an intuitive interpretation that smoother excision trajectories must result in a more uniform application of excision forces. The smoothness of the excision forces achieved by two medical students (dotted green lines) and two practising surgeons (dashed red lines) suggest that more experienced surgeons are likely to apply more uniform blade trajectories.

Fig. \ref{fig:surgeons-trainees} (middle) shows the contour plots of the Amplitude and Consistency features of excision forces against the sawing parameter $\rho$. The results show that $\rho$ has no significant correlation with the excision force amplitude - both smooth and sawing excision trajectories can yield equally low or high cutting forces. On the other hand, the consistency of excision forces (a feature that reflects the inverse of the spread of force levels during excision), highlights a strong correlation with sawing parameter $\rho$ (Pearson's $r=0.82$, $p < 0.05$). This relationship is explained by Fig. \ref{fig:model_pred} (bottom row), where larger $\rho$ values correspond to the noticeably lower variations of $z_b$, and as the result, to lower variations along $x$, $y$ and pitch components of the excision trajectories. This observation agrees with a general intuition that it is more difficult to apply excision forces consistently when sawing. 

The Confidence feature (Fig. \ref{fig:surgeons-trainees} right), which characterises both the uniformity and consistency of the excision forces, shows a significant alignment with sawing parameter $\rho$ (Pearson's $r=0.78$, $p < 0.05$). Similar to the Amplitude feature, the experiment results show a weak relationship (Pearson's $r=0.58$, $p < 0.05$) between $\rho$ and the Energy feature. 
Finally, Fig. \ref{fig:surgeons-trainees} shows the model $\mathcal{F}$ characterisation of excision forces from two professional surgeons (denoted as \textbf{S}) and two medical trainees with no experience at elliptical excision task (denoted as \textbf{T}). Notice that model characterisation of surgeon excisions is aligned with the higher sawing parameter $\rho$ values, whereas the performance of trainees matches the region of lower $\rho$ values. Again, this indicates that surgeons are likely to exhibit smoother excision trajectories when compared to less experienced medical students. 

\section{Discussion and conclusions}
In this study, we used a model of excision forces to optimise the pose trajectories of a blade in an elliptical excision task.
We proposed an excision trajectory generator capable of producing human-like elliptical excision behaviours parameterised by a single parameter $\rho$ that defines the amount of sawing motion applied during excision. 
These models can be used to efficiently optimise the excision technique with desired excision characteristics using Bayesian optimisation.
More specifically, we show how to align robotic excision behaviours with features that satisfy the criteria of surgical experts and behaviours that resemble performance characteristics of surgeons at various experience levels.
Experimental results indicate that the proposed $\rho$-parameterisation can successfully produce cutting force modulation that maximises the performance assessment of experts.

Our analysis suggests that professional surgeons are more likely to apply smoother excision trajectories, whereas less experienced medical trainees exhibit greater sawing behaviour. 
It is well known that sawing movement assists the cutting process by lowering the forces required to separate the material \cite{atkins2004cutting}. 
We hypothesise that inexperienced trainees employ this strategy to overcome the need to apply excessive excision forces to perform a controlled cut.
Although this study is limited to scalpel trajectories only, the cutting forces are strongly affected by the tissue tensioning controlled by the non-dominant hand. 
Hence, we hypothesise that experienced surgeons, while applying the smooth excision trajectories, actively tension the tissues to assist the excision. This combination of excision and tissue tensioning strategies is an interesting line of future work to better understand the dexterous manipulation skills underlying the elliptical excision task. 
Finally, we want to emphasise the promising prospect of using the proposed framework for studying surgical techniques at various levels of expertise, analysing common mistakes and their effects on the tissues.

\bibliographystyle{IEEEtran}
\bibliography{references.bib}

\end{document}